\documentclass[preprint,review,12pt]{elsarticle}
\usepackage[vmargin=2cm]{geometry}
\usepackage{multirow}
\usepackage{float}
\floatstyle{plaintop}
\restylefloat{table}
\restylefloat{figure}
\usepackage{multirow}
\usepackage{pdflscape}
\usepackage{xcolor}
\usepackage{hyperref} 

\usepackage{graphicx}
\usepackage{caption}
\usepackage{subcaption}
\usepackage{amssymb}
\usepackage{amsmath,}
\usepackage{physics}
\usepackage{algorithm}
\usepackage{algorithmic}
\usepackage{longtable}

\journal{Elsevier}

\begin{document}

\begin{frontmatter}


\title{Heterogeneous Oblique Double Random Forest}



 \author[label1]{M.A. Ganaie}
 \ead{mudasirg@umich.edu}
\author[label2]{M. Tanveer\corref{cor1}}
\ead{mtanveer@iiti.ac.in}
 \author[label3]{I. Beheshti}
 \ead{beheshtiiman@gmail.com}
 \author[label4]{N. Ahmad}
\ead{nehalahmad@iiti.ac.in}
\author[label5]{P.N. Suganthan}
 \ead{p.n.suganthan@qu.edu.qa}
\cortext[cor1]{Corresponding author}

\address[label1]{Department of Robotics, University of Michigan, Ann Arbor, MI, USA} 
\address[label2]{Department of Mathematics, Indian Institute of Technology Indore, Simrol, Indore, 453552, India}
\address[label3]{Department of Human Anatomy and Cell Science, Rady Faculty of Health Sciences, Max Rady College of Medicine,University of Manitoba, Winnipeg, MB, Canada\fnref{label4}}
 \address[label4]{Department of International Program of Electrical Engineering and Computer Science, National Taipei University of Technology, Taipei, Taiwan}
\address[label5]{KINDI Center for Computing Research, College of Engineering, Qatar University, Doha, Qatar \vspace{-1cm}}

\begin{abstract}
The decision tree ensembles use a single data feature at each node for splitting the data. However, splitting in this manner may fail to capture the geometric properties of the data. Thus, oblique decision trees generate the oblique hyperplane for splitting the data at each non-leaf node. Oblique decision trees capture the geometric properties of the data and hence, show better generalization. The performance of the oblique decision trees depends on the way oblique hyperplanes are generate and the data used for the generation of those hyperplanes. Recently, multiple classifiers have been used in a heterogeneous random forest (RaF) classifier, however, it fails to generate the trees of proper depth. Moreover, double RaF studies highlighted that larger trees can be generated via bootstrapping the data at each non-leaf node and splitting the original data instead of the bootstrapped data recently. The study of heterogeneous RaF lacks the generation of larger trees while as the double RaF based model fails to take over the geometric characteristics of the data. To address these shortcomings, we propose heterogeneous oblique double RaF. The proposed model employs several linear classifiers at each non-leaf node on the bootstrapped data and splits the original data based on the optimal linear classifier.  The optimal hyperplane corresponds to the models based on the optimized impurity criterion. The experimental analysis indicates that the performance of the introduced heterogeneous double random forest is comparatively better than the baseline models.  To demonstrate the effectiveness of the proposed heterogeneous double random forest, we used it for the diagnosis of Schizophrenia disease. The proposed model predicted the disease more accurately compared to the baseline models.
\end{abstract}


\end{frontmatter}

\section{Introduction}
The performance of classifiers can be enhanced by deploying multiple independent parallel classifiers. The process of deploying multiple classifiers together and combining their output to make the final decision of the model in such way that the generalization performance of the model increases. The pivot approach of ensemble neural networks is the perturb and combine strategy \cite{breiman1996bias, dietterich2000ensemble}. This approach is widely used in various machine learning specific domains \cite{wiering2008ensemble} such as computer vision (CV) \cite{goerss2000tropical} and recognition of various meshed surface patterns such as geometric pattern and semi-regular patterns. Models having multiple classifiers \cite{zhou2013multiple} or the learning based on ensemble perturb the original input training data to generate more diversified data samples among ensemble neural network base learners. It utilizes an aggregation strategy to combine outputs of base learners inorder to achieve better generalizability of ensemble network than an individual network. The literature of ensemble classifiers witness that its performance is better than an individual classifier \cite{dietterich2000ensemble}.

The superiority of ensemble learning models over individual learning models is that it has better prediction accuracy and it reduces the variance among the individual base learning models \cite{breiman1996bias, geurts2006extremely}. In the theory of variance reduction \cite{breiman1996bias, kohavi1996bias}, the classification error is expressed in terms of bias and variance. The bias metric measurement expresses the deviation between the target output class and the average predicted output of each individual base learners over the perturb training datasets generated from the input training  dataset. The variance measurement expresses the amount that base learners predictions fluctuate with the noise perplexities of the input training data. 

The most commonly utilized data classification technique is the decision tree due to its capability of more simplicity, integrity and powerful capacity of interpretability. To partition the input data recursively, decision tree utilizes divide and conquer method. Partitioning the data recursively by the decision tree is more sensitive to perplex the input samples, this lead the classifier away from stabili. Therefore, it is considered to keep the variance high and the bias low. Inorder to enhance the classification accuracy of the unstable data classifiers, ensemble algorithms are utilized.

The eminent classifiers which are more commonly used are rotation forest \cite{rodriguez2006rotation} and random forest \cite{breiman2001random}. Fundamentally these two classifiers are based up on ensemble approach and uses decision tree and their basic data classifier. Researchers in  \cite{ganaie2022oblique} expressed random forest as the best classifier among different $179$ classifiers experimented on the comprehensive and large scale $121$ datasets due to its better capability of generalization \cite{fernandez2014we}. 

 The decision tree's ensemble technique, random forest \cite{breiman2001random} utilizes bagging technique \cite{breiman1996bagging} and random subspace algorithm \cite{ ho1998random}. The base learners receive a wide diversity by these two major approaches for better generalization such as decision tree. Bagging, called as bootstrap aggregation technique is capable of generating plentiful bags of input training dataset in such a way that each bag  of data can be used to train each decision tree algorithm. Every decision tree is trained with a bag of dataset whose distribution is analogous to the total number data points, therefore, each classification algorithm express better generalization model performance.
 
 
 Inorder to achieve an enhanced generalization performance, diversified hyperparameters of random forest algorithm are  optimally selected. One of these hyperparameters is the decision tree (a base learner) in the random forest (ntree), the count of candidate features to be evaluated at any nonleaf (mtry) node, and the count of impure node samples (node size or minleaf).
 
Several studies have been proposed so far such as tuning process analysis \cite{probst2017tune}, parameter sensitivity \cite{huang2016parameter}, ramification of tree counts in ensembles \cite{banfield2006comparison, hernandez2013large} gives acumen how the parameters influence the performance of the model. To retrieve as many potential candidate features as feasible, various algorithms \cite{boulesteix2012overview, han2019optimal} have been introduced. 
The hyperparameters are better chosen after analysing the sample size optimally in bagging \cite{martinez2010out} and tree size estimation using a consolidation of random forest and adaptive nearest neighbours \cite{lin2006random}.

In general, two approaches are used to generate decision trees, univariate \cite{banfield2006comparison} and multivariate \cite{murthy1996growing} decision trees. The univariate decision tree is called as the parallel axis or orthogonal tree, utilize multiple impurity criteria for the optimization of the best feature univariate split from multiple randomly selected subspaces of features. Decision trees of multivariate type synonymously called as oblique decision tree (ODT),  operate by splitting the tree nodes using multiple features.
Generally, the decision boundary can be approximated by multiple stair-like decision boundaries of an  ODT.
Random forest algorithm is a univariate decision tree, it generates non-terminal hyperplane nodes in such a way that bifurcating of children nodes of any decision tree becomes easier. 
At any non leaf node of a decision tree, it is not considered as a good classifier in case of splitting the hyperplane \cite{manwani2011geometric}. Inorder to determine the best split algorithm among the split sets, different conditions such as Gini index measure, entropy measure and towing rules are ramified in most of the models based on decision tree in such a way that the impurity score remains lowest after split. The distribution skewness of various category samples are measured from the impurity criteria at every non-leaf node. 
Low impurity score is determined at approximately homogeneous distribution and high impurity score is assigned to the distribution where any class is dominated by another specific class. Part of the impurity measure is optimized for tree generation in maximum induction tree algorithms which are based up on decision tree. As the impurity measurements are non differentiable as compared to the hyperplane parameters, the decision trees are generated using various search algorithms. Such as deterministic hill-climbing algorithm of CART-CL \cite{breiman2017classification}, randomized search based CART-LC in OCI \cite{murthy1994system}. If the feature space is of high dimensional, these two algorithms suffer due to one dimensional search and the problem of local optima. Thus multiple iterations are utilized to avoid the problem of local optima. In addition, several evolutionary algorithms have been applied for multidimensional optimization \cite{pedrycz2005genetically, cha2009genetic} which can endure noisy rating function evaluation while also optimizing the numerous rating functions \cite{cantu2003inducing, pangilinan2011pareto}. Trees randomized extremely \cite{geurts2006extremely} and the oblique variant \cite{zhang2014towards}, firmly randomized the cut points and attribute sets. Some other methods include decision trees based on fuzzy inference systems \cite{wang2008induction, wang2008improving}, feature space \cite{zhang2014random} ensemble and support vector machine based decision trees \cite{zhang2007decision}. Weights of random feature  for ensemble decision tree assign weight to each individual attribute for better model variation \cite{maudes2012random}. The interpretability of featured decision trees has been studied recently inorder to better understand how they make decisions \cite{sagi2020explainable, fernandez2020random}. More literature information about decision trees can be studied from \cite{rokach2016decision}. 

The problem with all of the impurity measures, as mentioned in \cite{manwani2011geometric}, is that they are different class distribution functions of the hyperplane on both of its sides and disregard the spatial structure of the regions of its class \cite{manwani2011geometric} and if the labels of the data are changed without any modification in the significant features of  hyperplane edges, it has no effect on the impurity metric. 
Support vector machines are utilized as decision tree generator to incorporate the spatial structure of featured class distribution \cite{manwani2011geometric}. The proximal hyperplanes generated by multisurface proximal support vector machines (MPSVM) are in such a way that each proximal hyperplane is closer to the class samples and far away from another class samples \cite{mangasarian2005multisurface}. Inorder to make the node pure, authors \cite{manwani2011geometric} produced planes of two clustering at non-leaf node level by selecting an angle bisector. MPSVM algorithm is considered in the binary class. Therefore, the multiclass problem has been split into binary class considering superior class samples in one group and  remaining  into another class sample. As the tree grows bigger, the nodes reach towards its purest form and the subsequent nodes receives less samples. To avoid this issue, NULL space algorithm \cite{chen2000new} is deployed \cite{manwani2011geometric}. Similarly, regularization algorithms such as axis-parallel split regularization and Tikhonov regularization \cite{marroquin1987probabilistic} were deployed in the oblique decision tree ensemble based up on MPSVM algorithm \cite{zhang2014oblique}. The twin bounded SVM \cite{shao2011improvements} produced better performance in terms of model generalization due to the fact that it does not require explicit generalization technique to overcome these problems \cite{ganaie2020oblique}. Inorder to find the optimal node split out of the split node candidates, both ensemble oblique decision tree based on MPSVM and ensemble oblique decision tree based on TBSVM utilize exclusive base learner of an individual nonleaf node. Generalization performance of ODT ensemble express better results than classic random forest algorithm \cite{zhang2017benchmarking}.

Linear discriminant analysis, MPSVM, logistic regression, ridge regression and least squares SVM  are used to produce hyperplanes in a heterogeneous oblique random forest \cite{katuwal2020heterogeneous}. Those generated hyperplanes which produce purer nodes are selected as the best split optimal hyperplane. The node size impact on the model performance evaluated recently in several studies of double random forest(DRaF) \cite{han2020double}. According to the findings of the study, it was revealed that the model prediction performance is directly proportional to the depth of the decision tree, deeper the generated decision tree better the performance of the model. The authors also brought to light that the largest tree generated by the classical random forest algorithm on a set of data may not be competently large enough for the optimum model performance. Hence, the decision trees generated by  DRaF algorithms \cite{han2020double} are much better and bigger than the one generated by classical random forest algorithms. Instead of getting the decision tree trained each using distinct bags of training datasets generated at root node by bagging strategy, the authors \cite{han2020double} constructed each tree with the original dataset available for training and to obtain the optimal split node, bootstrap aggregation is utilized at individual non-terminal nodes of the decision tree. The random forest algorithm as well as the DRaF algorithm are univariate, therefore, ignores the distribution of the class geometrically which results the generalization performance very low. Moreover, the heterogeneous DRaF uses bootstrapped data for only at the root node and split the boostrapped data which results in generation of trees with lower depth \cite{han2020double}.  To address the above issues, a heterogeneous oblique double random forest (Het-DRaF) has been proposed. The proposed Het-DRaF uses multiple linear classifiers on non-leaf nodes and selects the hyperplane which optimizes the impurity criteria.
The proposed Het-DRaF model is different from the existing classifiers as follows:
\begin{itemize}
\item Unlike the standard RaF algorithm and heterogeneous oblique RaF which use bootstrapping at the root node, the proposed Het-DRaF model use bootstrapping on non-leaf nodes.
    \item 
Unlike the standard RaF algorithm and heterogeneous oblique random forest, the proposed Het-DRaF model optimizes the splitting hyperplane based on the bootstrapped data on non-leaf nodes and send the original training input data rather than the bootstrapped data. 
\item Unlike standard DRaF and standard RaF which are univariate decision trees and fail to capture the geometrical characteristics of the data, the proposed Het-DRaF model generates multivariate decision trees to take over the geometric properties of the data.
\item Unlike the standard oblique decision trees based ensembles (MPRaF-P, MPRaF-T, MPRaF-N) which use a single decision hyperplane on non-leaf nodes based on the bootstrapped data, the proposed Het-DRaF model uses multiple linear classifiers to generate the best split based on the original data rather than bootstrapped data. 
\end{itemize}

The remaining part of the article is organised as: Section \ref{sec:2} presents the literature review part of decision tree ensemble algorithms. Section \ref{sec:3} presents the proposed Heterogeneous Double Random Forest algorithm. Section \ref{sec:4} gives the brief explanation about experimental analysis on UCI datasets. Section \ref{sec:5} gives the brief explanation about experimental analysis on schizophrenia dataset followed by the conclusion and future work describes in section \ref{sec:6}.

\section{Related works}
\label{sec:2}
 In this part, we briefly discuss the literature review of decision trees ensembles.
 \subsection{Multiclass problems and solutions}
Several popular binary class classifiers utilize the common approach of 
``one-vs-all" for the purpose of breaking down the multiclass  problem into multiple binary classes. In this approach, multiple binary classifiers are constructed and trained to differentiate the training set into a unique class from the all the remaining classes. Inorder to classify a  new dataset, multiple classifiers deployed and the one that gives highest confidential output is selected. Inorder to find the classical separation of a node (non-terminal) hyperplanes of any decision tree that can be considered as the multiclass problem because MPSVM model produces binary classification of data. Therefore, to overcome the multiclass problem by using binary class algorithms is a heavy task and there are many different algorithms available such as one-versus-one \cite{knerr1990single}, one-versus-all \cite{bottou1994comparison}, error correcting output codes \cite{dietterich1994solving}, decision directed acyclic graph \cite{platt1999large} and many other have been proposed so far. Data splitting strategy of the non-leaf node decision trees has better preference over different binary classification algorithms \cite{zhang2014oblique}. Separation of majority data samples into one category and  remaining samples into another category consequences to an ineffective general algorithm because it declines to occupy the spatial structure of samples of the data \cite{manwani2011geometric}. To assimilate the spatial structure, authors \cite{zhang2014oblique} moldered the problem of multiclass classification by utilizing the separability information into a binary class problem. The researchers \cite{zhang2014oblique} utilized Bhattacharyya distance for the assimilation. Statistically, Bhattacharyya distance provides the similarity measurement between the two continuous or discrete probability distributions as it is assumed to offer an excellent insight into the separability between two ordinary classes:
$
K_{1} \sim N\left(\gamma_{1}, \mu_{1}\right) 
$,$
K_{2} \sim N\left(\gamma_{2}, \mu_{2}\right).
$
  where $\gamma_{i}$ and $\mu_{i}$ represents the normal distribution class parameters of class $K_i$, for $i = {1, 2}$. Considering similar strategy \cite{zhang2014oblique}, in this research we utilized multivariate Gaussian distribution \cite{jiang2011linear}. We employ Bhattacharyya distance for the measurement of class separability inorder to decompose a multiclass issue into a binary problem. \par
  
\begin{table}[h]
\label{algo:1}
\hrule 
\textbf{Algorithm 1:} Formation of binary class problem from a multiclass problem 
\hrule  
\textbf{Input:} \\
$X=\mathrm{N} \times f$ is the dataset with $N$ data points and a feature size of $f$. \\
$\mathrm{Z}=\mathrm{N} \times 1$ is the model output. \\
${ {\beta_1}}, \beta_2, \ldots, {\beta}_C$ target values.\\
\textbf{Output:} \\
$G_{p}$ and $G_f$ are the hyperclasses or clusters,  \\
For $j=1,2,\cdots,C.$ 
\begin{enumerate}
    \item For a class pair  $\beta_{j}$ and $\beta_{k}$, with $k=j+1,\cdots, G$, the Bhattacharya distance can be calculated  as:
$F\left(\beta_{j}, \beta_{k}\right)=\frac{1}{8}\left(\eta_{k}-\eta_{j}\right)^{t}\left(\frac{\sum_{j}+\sum_{k}}{2}\right)^{-1}\left(\eta_{k}-\eta_{j}\right)+\frac{1}{2} \ln \frac{\mid\left(\sum_{j}+\sum_{k}\right) / 2  |}{\sqrt{|\sum_{j}||\sum_{k}|}}.
$
Here,  $\eta$ is the mean and $\sum$ is the covariance at $j^{th}$ and $k^{th}$ cluster, respectively.
    \item Obtain the pair $\beta_{p}$ and $\beta_{f}$ of clusters with the highest distance (Bhattacharyya) and assign to $G_{p}$ and  $G_{f}$.
    \item for other clusters, if $F\left(\beta_{k}, \beta_{p}\right)<F\left(\beta_{k}, \beta_{n}\right)$ then put $\beta_{k}$ to $G_{p}$ Otherwise put in $G_f$.  \hrule 
\end{enumerate}

\end{table}

\begin{table}[h]
\label{algo:2}
\hrule
\textbf{Algorithm 2:}  RaF (Random Forest)\
\hrule
\textbf{Training:}\\
$X:=N \times f$ be input data sample of $f$ features in $N$ data points.\\
$Z:=N \times 1$ is target value.
$\beta$ : is the number of base learners.\\
``mtry": is the candidate feature to be computed non-leaf node level.\\
``nodesize" or ``minleaf": amount of data can be put into the impure node.\\
For a DT, $T_{i}$ (for  $i=1,2, \ldots, \beta$)
\begin{enumerate}
    \item Bootstrap data samples generation $X_{i}$ from $X$.
    \item DT generation using $X_{i}$ :\\
    For any node $d$ :
    \begin{enumerate}
        \item  Select ``mtry" $=\sqrt{f}$ features from feature space of $X_{i}$ .
        \item  Chose the feature with the best split from the $X_{d}^{*}$ random feature subset  along having the cut points.
        \item  Using the the cut point and optimal split feature, divide the samples of data.\end{enumerate}
Reiterate steps (a)-(c) until met the stopping criteria.
\end{enumerate}
\textbf{Classification:}\\
For any data $x_{i}$, utilize base learners of the decision tree for data sample label generation.
The class prediction is obtained by majority voting.
\hrule
\end{table}
\begin{table}[]
\label{algo:3}
\hrule
\textbf{Algorithm 3:} DRaF (Double Random Forest)
\hrule
\textbf{Training}\\
$X:=N \times f$ , be the set of data samples with $N$ data points and $f$ size of features.
$X_{i}:=N_{i} \times f_{i}$ denotes the training data, received by $i^{th}$ node, with total number of samples $N_{i}$ and  $f_{i}$ feature size.\\
$Z:=N \times 1$ is the target value.\\
$\beta$ : is the base learners.\\
``mtry": is the candidate features to be computed on non-leaf nodes.\\
``nodesize" or ``minleaf": amount of data can be put into the impure node.\\
For a DT, $T_{i}$ for $i=1, 2, \ldots, \beta$
\begin{enumerate}
    \item Get input data $X$.
    \item Generate DT $T_{i}$ with random set of features and randomised based bootstrap instance utilizing $X$ :\\
          For any node $d$ having data samples $X_{d}$ :
        \begin{enumerate}
            \item if $N_{d}$ is greater than $N\times 0.1$\\
                produce bootstrap sample $X_{d}^{*}$ from $X_{d}$.\\
                  else\\
                    $X_{d}^{*}=X_{d}$
            \item Select ``mtry" $=\sqrt{f}$ features from feature space of $X_{d}^{*}$.
            \item Chose the best split from the random feature subset $X_{d}^{*}$ along with the cut points.
            \item Obtaining optimal split feature with $X_{d}^{*}$ along with cut point, split the data $X_{d}$ into leaf nodes.\end{enumerate}
         Reiterate step (a)-(c), until any condition below is satisfied:
         \begin{itemize}
             \item Purest form of a node is obtained.
             \item $minleaf$ is less or equal to the number of samples reaching a given node.
         \end{itemize}
    
\end{enumerate}
\textbf{Classification Phase:}\\
For any point of test data $x_{i}$, utilize the decision trees to obtain the test sample label.
The class prediction is obtained by majority voting.
\hrule
\end{table}
\begin{table}[]
\label{algo:4}
  \hrule 
\textbf{Algorithm 4:} ODRaF-MPSVM  \hrule 
\textbf{Training Phase}\\
$X:=N \times f$ is the input data set with total number of $N$ samples and feature size $f$.\\
$X_{i}:=N_{i} \times f_{i}$ is the input training data samples received by node $i$, having samples $N_{i}$ and feature size $f_{i}$.\\
$Z:=N \times 1$ is output labels.\\
$\beta:$ is the base learner.\\
``mtry": represents features computed over non-leaf nodes.\\
``nodesize" or ``minleaf": data that can be put into an impure node.\\
For a decision tree, $T_{i}$ (for $i=1,2, \ldots, \beta$)
\begin{enumerate}
    \item Get training data sample $X$.
    \item Develop DT $T_{i}$ with random set of features and bootstrap instance (randomised) utilising $X$.
    At any node $d$ of a sample $X_{d}$ :
    \begin{enumerate}
        \item    if $N_{d}$ is greater than $N \times 0.1$\\
         Generate samples $X_{d}^{*}$ from $X_{d}$.\\
          else \\  $X_{d}^{*}=X_{d}$
        \begin{enumerate}
            \item Choose ``mtry" $=\sqrt{f}$ from $X_{d}^{*}$.
            \item Deploying [Algorithm-1] cluster the data samples $X_{d}^{*}$ into $G_{p}$ and $G_{f}$.
            \item Obtain the best split using MPSVM (variation in regularization) with with $G_{p}$ and $G_{f}$ and split $X_{d}$ samples into leaf nodes.
        \end{enumerate}
Reiterate (i)-(iii), until stopping criteria mets:\end{enumerate}
\begin{itemize}
            \item Purest form of a node is obtained.
            \item $minleaf$ is less or equal to the number of samples reaching a given node
        \end{itemize}
\end{enumerate}
Classification Phase:
For any point of test data $x_{i}$, utilize the decision trees to obtain the labels of test data.
The class prediction of the test data point is obtained by the majority voting.
\hrule
\end{table}
\begin{table}[h!]
\label{algo:5}
\hrule
\textbf {Algorithm 5:} DRaF with PCA  \hrule
{\text { Training }}$ \\
X:=N \times n$ be the set of data samples with $N$ data points and $n$ size of features.
$X_{t}:=N_{i} \times f_{i}$ is the training samples received at $i_{th}$ node, having $N_{i}$ samples with $f_i$ as feature size.\\
$Z:=N \times 1$ is the target value.\\
$\beta$ : is the base learners\\
``mtry": represents features computed over non-leaf nodes.\\
``nodesize" or ``minleaf": amount of data to be put into an impure node.\\
For a DT , $T_{i}$ for $i=1,2, \ldots, \beta$
\begin{enumerate}
    \item Get input samples for training $X$.
    \item Obtain the DT $T_{i}$ with random set of features and bootstrap instance (randomised) utilising $X$:\\
    For a node $d$ having data sample $X_{d}$.
    \begin{enumerate}
        \item if $X_{d}$ is greater than $X\times 0.1$
        \item Generate samples $X_{d}^{*}$ from $X_{d}$.
        $X_{d}^{*}=X_{d}$
\begin{enumerate}
    \item   Choose ``mtry" $=\sqrt{n}$ from $X_{d}^{*}$.
    \item Evaluate the scatter matrix $S_{d}$ by $X_{d}$.
    \item Evaluate eigenvectors ($V$) of $S_{d}$.
    \item Evaluate transformation of data employing eigenvectors $V$ as, $X_{P C A}^{*}=X_{d}^{*} * V$.
\end{enumerate}\end{enumerate}Repeat (i)-(iii), until met any one the stopping criteria.\end{enumerate}
\textbf{Classification:}\\
For an input $x_{i}$, obtain labels using DTs. The class prediction is computed by majority vote.
\hrule
\end{table}
\subsection{Random Forest}
Bagging and random subspace algorithms are fundamentally utilized to generate decision trees ensemble known as random forest algorithm \cite{breiman2001random}. Here, decision tree is utilized as a base learner. The diversified ensemble decision trees are produced by the two renowned approaches random subspace and bagging. At each node Optimal split is chosen by the ensemble decision tree among the subsets of total splits. The best split is determined by employing an impurity criterion such as Gini impurity, information gain and several others \cite{breiman2017classification}. The classical RaF algorithm is represented in Algorithm-$2$. The CART (classification and regression tree) can execute test split by utilizing only a single feature. Therefore, it is also called as the univariate decision tree \cite{murthy1996growing}. 

\subsection{Double random forest(DRaF)}
An another ensemble of DT is the DRaF \cite{han2020double} which is also based on the basic concept of random subspace algorithm and bagging. Unlike classical random forest algorithms wherein the training of base learner is accomplished by using the bootstrapped data samples, the DRaF algorithm utilizes original dataset to train its base learners, resulting exclusively more unique samples in the training of DRaF than classical random forest. As the uniqueness of instances in training increases, the decision tree becomes larger and gives enhanced generalization performance. The bootstrap sampling approach is employed momentarily in the DRaF algorithm at each non-terminal position of nodes. After selecting the feature that results in splitting  from a randomly selected features from the data samples of bootstrapping , once the final splitting of input dataset is completed then this data is fed to bottom levels of decision tree.  Algorithm-$3$ expresses the DRaF. 

\subsection{Oblique ensembles of double random forest (ODRaF) with MPSVM}
Generation of DRaF ensembles is called oblique ensembles of DRaF. Geometrically, decision trees of univariate type do not take into account general input data attributes. Both DRaF algorithm and classical random forest algorithms come in the category of univariate ensemble decision trees. Also, the classical random forest based decision tress are not large enough to achieve the better generalization performance for the dataset. These limitations can be addressed by ODRaF along multisurface proximal SVM (MPSVM) \cite{Ganaie2022EnsembleDL, ganaie2022oblique}. Unlike classical random forest algorithms, bootstrapping samples are used in ODRaF algorithms with MPSVM on a non-terminal node (unless a set of finite criteria is full filled as represented in Algo-$4$) for optimal oblique splits generation and then instead of bootstrapped samples division, the original data is divided among the leaf nodes. Inorder to assimilate the spatial shape in the hyperplane which is splitting, authors introduced ODRaF with MPSVM \cite{Ganaie2022EnsembleDL, ganaie2022oblique} in which the MPSVM generates optimal split at non-leaf nodes. As the dimension of the decision tree grows bigger, the number of data points arriving towards any specific node decrease, resulting in problem of sample size variation. Inorder to address this problem, authors \cite{Ganaie2022EnsembleDL} utilized different techniques for obtaining regularization performance. When a model utilize Tikhonov regularization algorithm, then this is called as ``oblique DRaF via MPSVM with Tikhonov regularization" (MPDRaF-T) on the other hand, if it employs parallel axis split regularization then it is known as ODRaF via MPSVM with parallel axis split regularization (MPDRaF-P) \cite{Ganaie2022EnsembleDL}. Inorder to regularize the data samples, the summation of positive integer is done along the diagonal elements present in the matrix may be $(Y)$ that is, if r is rank deficient matrix of  $Y$, then regularize it as:
$
    Y = Y + \delta \times I;
$
here, $\delta$ is a positive integer, $I$ is the corresponding identity matrix of appropriate dimension. A decision tree is being grown to completion when the sample matrix $Y$ at any node is rank deficient under split regularization. Hence, for the decision tree growth heterogeneous functions are utilized. The ODRaF with MPSVM is summarized in Algorithm-$4$.

\subsection{DRaF with PCA/LDA}
Rotation based DRaF ensemble algorithms are employed to generate the ensemble divers learners.
Distinct projections emerge from rotation of various random feature subspaces, resulting in an improved generalisation model performance. The motive of this method is to transform the input sets of data which are more diversified between the learners. Inorder to achieve more diversified base classifiers, random feature subspace transformation is applied at non-leaf nodes. The transformation is accomplished by the two most popularly used principals namely linear discriminant analysis (LDA) and principal component analysis (PCA). The DRaF model with PCA (DRaF-PCA) can be seen in Algorithm-$5$. The transformation is performed on bootstrapped samples at every non-leaf nodes and moving towards a node with random feature space. \par
The DRaF with LDA (DRaF-LDA) differs from DRaF model with PCA (DRaF-PCA) in second and third step. In DRaF-LDA algorithm, rather than computing node level scatter matrix $S_d$, it computes scatter matrix $S_d^w$ and the in between class scatter matrix $S_d^b$. Then the eigenvector of ($S_d^w$, $S_d^b$) are computed as 
$
 S_d^b \times \alpha = \lambda \times S_d^w
$
here, $\alpha$ is an eigen vector and $\lambda$ is eigen value, respectively.

\section{Proposed Heterogeneous Double Random Forest}
\label{sec:3}
Random forest \cite{breiman2001random} and DRaF \cite{han2020double} have been successfully employed in the classification problems. However, both RaF and DRaF models generate the axis parallel splits based on a single feature from the given feature space, hence, known as univariate decision trees. Univariate decision trees neglect the spatial data structure. To take the geometric properties of the data in account, oblique random forest \cite{zhang2014oblique} generated the decision trees based on the MPSVM. MPSVM generates the clustering hyperplane at each node for splitting the data. Generating the hyperplanes at each node via single family classifier may not always create the best partition or capture the geometric property properly as the decision boundary at a given node depends upon the data reaching to that node. Moreover, the random forest \cite{breiman2001random}, oblique random forest and heterogeneous random forest \cite{katuwal2020heterogeneous} may not generate the decision trees of sufficient depth \cite{han2020double, Ganaie2022EnsembleDL} as these models generate the decision boundaries based on the bootstrapped data which may posses fewer unique instances. To address these problems, a heterogeneous oblique double random forest (Het-DRaF) has been proposed for the classification. Het-DRaF uses whole training data to generate the decision trees unlike the standard random forest based models which use bootstrapped data for generating the decision trees. Hence, the proposed Het-DRaF model use the same training data to generate all the decision trees whereas the standard random forest based models utilise the bootstrapped data to generate the trees which has fewer unique instances.

Het-DRaF model uses several linear classifiers from different families at each non-leaf node to generate the best decision split. We employed ridge regression (RR), linear regression (LR), support vector machine (SVM), linear discriminant analysis (LDA),  multisurface proximal support vector machine (MPSVM) and least square support vector machine (LSSVM) to generate the decision plane at each node. Although a single classifier may not be always best, however, we can integrate the several linear classifiers for generating the best split in the Het-DRaF. Based on the \cite{zhang2014oblique}, for a $K$ class problem, $nK$ evaluations are performed at each non-leaf where $n$ represents the number of linear classifiers. Theoretically, highly balanced decision trees are desirable \cite{millington2018artificial} but in practice it is very hard to guarantee the balanced trees. For generating the optimal hyperplanes, the partitioning of the data at each node is of utmost importance. Mostly, one-vs-all approach is used  for partitioning  the data but it limits the generalizability as it leads to unbalanced trees and  also the search space is limited (only $K$ issues). Hence, we follow the strategy given in \cite{katuwal2020heterogeneous} and rank the partition splitting based on the Gini index. In multiclass problems, the oblique decision trees generate the splits for a binary class problem. Hence, the generalizability of the model is a function of splitting the data at each node following the training of classifiers. For splitting the data at each node, we use one-vs-all approach of \cite{zhang2014oblique} and partitioning approach based on \cite{zhang2014oblique,truong2009fast}. Authors in \cite{zhang2014oblique} partitioned  the data based on Bhattacharyya distance while as used all the possible choices of the partitioning \cite{truong2009fast}. 
At a given non-leaf node of a decision tree, the number of possible partitions is $2^{K-1}-1$ for a $K$ class problem \cite{truong2009fast}. As the number of classes increase, the partitioning of the data poses a computational issue. Hence,  we follow \cite{katuwal2020heterogeneous} to find the best partitions based on the Ideal Gini score and cluster separability to rank the partitions. We use genetic algorithm if the number of classes is greater than eight. Thus, based on the aforementioned points we partition the data at a given node.
Based on the partitions generated, we employ RR, LR, LDA, SVM, MPSVM and LSSVM classifiers to obtain the best split. As the best split is found corresponding to the given data partition, we recover the original data in the node and send the instances down the node. Note that this is different form the standard random forest based models as it uses bootstrapped data only at the root node instead of the original data.

In summary, the proposed Het-DRaF model is more diverse as it employs multiple linear classifiers at each node to generate the best split. Moreover, we follow different strategies like GA, Ideal Gini Score and the cluster separability to obtain the permissible partitions data. Unlike the standard RaF models which employ bootstrapping at the root node only, the proposed model uses bootstrapped data on non-leaf nodes to obtain the best split while sending the input data down to the child nodes. This results in more unique samples and leads to more accurate results. 

\section{Experimental Analysis on UCI Datasets}
\label{sec:4}
In this section, experimental setup, the experimental analysis and performance of baseline models and proposed models is discussed in detail. Here, the baseline models are tandard  RaF \cite{breiman2001random}, MPRaF-T 
\cite{zhang2014oblique}, standard DRaF \cite{han2020double}, MPRaF-P \cite{han2020double}, RaF-PCA \cite{zhang2014random} MPRaF-N \cite{han2020double} and  RaF-LDA \cite{zhang2014random}.


 \subsection{Statistical Analysis}
\begin{figure}
\caption{Nemenyi post hoc test assessment of classification (significance level $\alpha = 5\%$). Statistically identical classification models are connected.}
\centerline{\includegraphics[width=6in, height=2in]{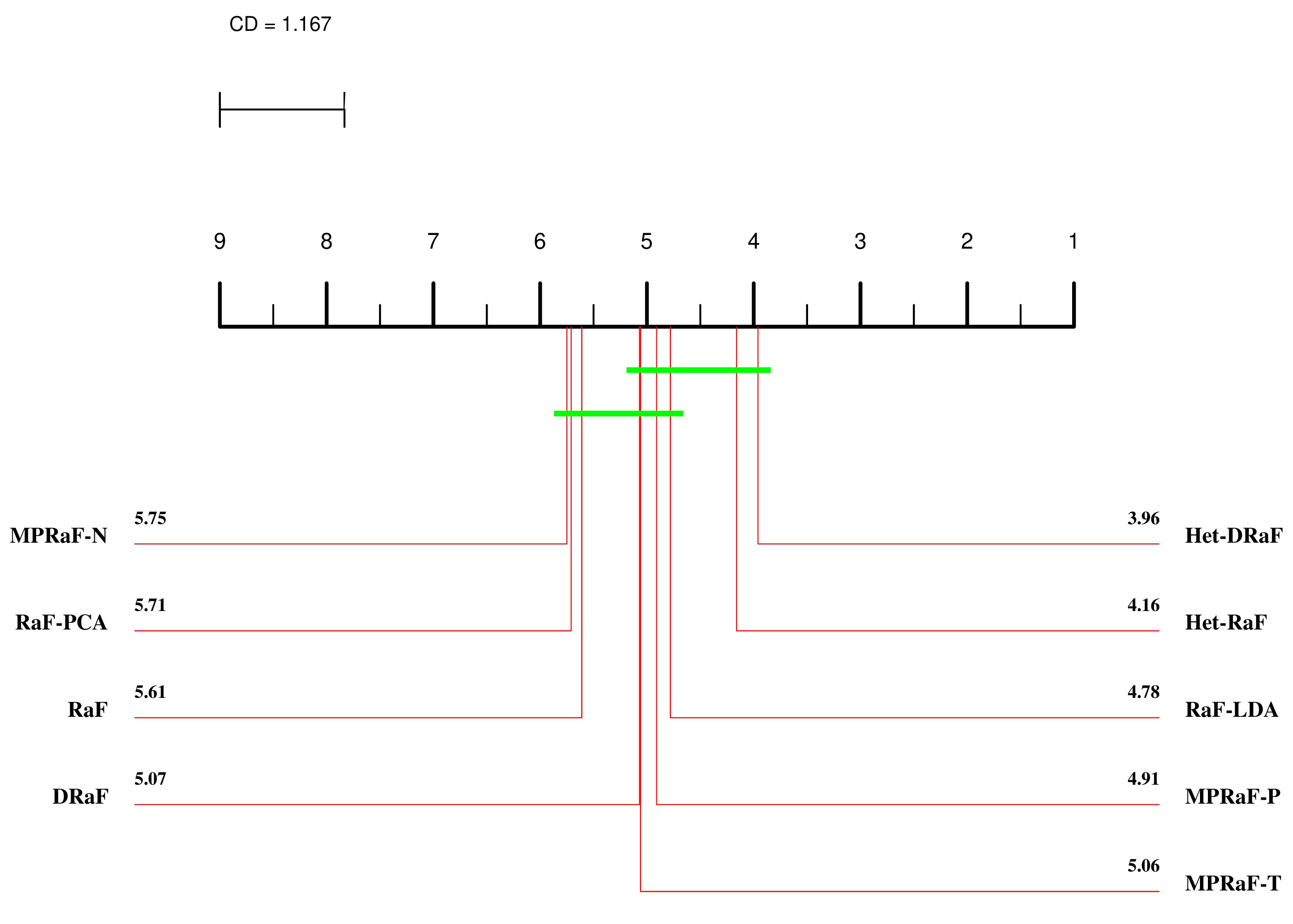}}
\label{fig:NemnyiTest}
\end{figure}

 Table \ref{table:averageRaF} represents classification performance summary of all the ensemble models used in the experiment on the $121$ datasets. It can be seen that the average accuracy (AA) of the proposed Het-DraF model outperformed the state-of-the-art model classifiers. According to the observations of \cite{fernandez2014we}, each classifiers are ranked based on their performances on $121$ datasets. The Friedman test ranks the classifiers based on their performance, lower performance model will get the higher rank conversely the same. Thus, low rank of the classifier model provides better generalization capability. Table \ref{tab:average Friedman Rank} represents the average rank of different classifiers. When comparing with state-of-the-art classifiers it is evident that the ensemble models Het-RaF and Het-DRaF provide outstanding classification performance. Moreover, the average rank of Het-RaF and Het-DRaF is lower than other existing classifiers.
 In this research, model performance is obtained by statistical analysis. Amongst the several hypothetical testing algorithms the most commonly used Friedman test \cite{demvsar2006statistical} is deployed in this experiment followed by corresponding Nemnyi post hoc test is utilized for model comparison. Let us consider the rank of $j^{th}$ classification model is $k^{j}_{i}$ on $i^{th}$ set of data, here, the total number of datasets are $N$. The average rank of the subjects $\sum_{i} r_{i}^{j}$ are evaluated using Friedman test for classification, when both number of classifiers $n$ and the total datasets $N$ are larger. The Friedman statistical analysis can be performed by the equation:
 $
\chi_{F}^{2}=\frac{12 N}{n(n+1)}\left[\sum_{j} R_{j}^{2}-\frac{n(n+1)^{2}}{4}\right]
$
where $\chi_{F}^{2}$ follows distribution of certain degrees of freedom $(n-1)$ in accordance with null hypothesis. Because $\chi_{F}^{2}$ is too conservative, the following equation can give more appropriate statistic
$
\label{equ}
F_{F}=\frac{(N-1) \chi_{F}^{2}}{N(n-1)-\chi_{F}^{2}}
$
which follows F-distribution with the classifier of the dataset $(n-1)\times(N-1)$ and $(n-1)$. It has been seen that all the classifiers behaves equal under the condition of null hypothesis, therefore, the ranks of all the classifiers are also same. In case the null-hypothesis If the null hypothesis does not hold true, then pairwise performance evaluation of the classifiers is given by the alternate Nemenyi post-hoc \cite{nemenyi1963distribution}. Two classifiers are said to be different if and only if the average rank of them is differ at least by the critical difference which is given by the following equation:
$
C D=q_{\alpha} \sqrt{\frac{n(n+1)}{6 N}}
$
where $q_{\alpha}$ is the statistic of standardized rank of the data and $\alpha$ is the significance level subdivided by $\sqrt{2}$.
Average ranks of classifiers RaF,  MPRaF-P, MPRaF-T, MPRaF-N,  RaF-LDA, RaF-PCA, DRaF, Het-RaF and Het-DRaF are $5.61, 4.91, 5.06, 5.75, 5.71, 5.07, 4.78, 4.16$ and $3.96$ respectively. 
And the average accuracy of the models  are $81.36, 81.55, 81.65,  80.55, 80.91, 81.55, 81.36, 81.94$ and $82.19$ respectively.
With general computation we obtained the value of $\chi^2_F=47.9247$, $F_F=6.2895$ at Significance level of $5$\%. F-distribution is followed by $F_F$ with the significant value of $N=106$, $K=9$ implies the value of $(n-1)=8$ and $(n-1)(N-1)=840$. From Table \ref{table:averageRaF} the Value of $F_F(8,840)=1.95$. As the value of $F_F$$>$$F_F(8,840)$ that is  $6.2895>1.95$, therefore, the null hypothesis is rejected. Hence it is validated that the classifier models has significantly difference among them. To obtain the exact difference value among the classification models, Nemenyi post-hoc test algorithm is utilized. By performing simple computation we obtain the critical difference(CD)$=1.17$ along with the value of $q_{\alpha=0.05} =3.1020$ at a significance level of 5\%. Looking into Fig. \ref{fig:NemnyiTest}, the existence of significant difference can be observed and these are unconnected through a line. Table \ref{tab:win-tie-loss} represents the summary of Nemenyi post-hoc algorithm results. From the results of  Table \ref{tab:win-tie-loss}, the performances of the proposed model and the state-of-the-art classification models cane be distinguished and it has been observed that the proposed models Het-RaF and Het-DRaF outperformed the state-of-the-art classification models such as RaF, MPRaF-P, MPRaF-T, MPRaF-N, RaF-LDA, RaF-PCA. 

\subsection{Sign test via win-tie-Loss }
Considering the null hypothesis, if every model wins the condition $N/2$ among $N$ datasets, then a pair of classifiers is said to be significantly different. Here, the binomial distribution is followed by the total number of wins. If the number of datasets (N) is too large, then the condition $N(N/2, \sqrt{N}/2)$ is followed by the number of wins, therefore, $z$-test is utilized here instead of any other categories of test. Two classification models are said to be better(if p \textless 0.05) if and only if the classifier model has the number of wins as $N/2+1.96\sqrt{N}/2$. Traditionally, null hypothesis is favored by tied matches, therefore the number of ties between the classifier models are split in even aspect while the odd numbers are ignored.
 It can be seen that the proposed Het-RaF and Het-DRaF algorithms grabs comparatively more wins than the stat-of-the-art models. Proposed Het-RaF and Het-DRaF algorithm outperformed and stand  winner among the MPRaF-N and RaFPCA existing models among $121$ datasets. When compared the proposed model to the baseline models, it emerged as the best performer in more datasets. Table \ref{tab:signifcdf} shows the classification performance among the baseline algorithms and the proposed Het-DRaF and Het-RaF with Nemenyi post hoc tests based on the accuracy and it represents that the proposed Het-RaF and Het-DRaF models outeperformed the existing algorithms RaF,MPRaF-P MPRaF-T, MPRaF-N, RaF-LDA, RaF-PCA and DRaF.

 \textbf{\section{Experimental Analysis on Schizophrenia Dataset}
\label{sec:5}}
The experimental results of the proposed model and the baseline models on Schizophrenia datasets are discussed in this section.

\subsection{Pre-processing of Schizophrenia Dataset}
The schizophrenia data were obtained from COBRE dataset (The Center for Biomedical Research Excellence in Brain Function and Mental Illness; \url{http://fcon_1000.projects.nitrc.org/indi/retro/cobre.html}, which included anatomical and resting-state functional MRI (rs-fMRI) scans from $72$ patients with Schizophrenia (mean age $\pm$ SD:$38.16$ years $\pm$ $13.89$, age range: $18–65$ years) and $74$ healthy controls (mean age$\pm $SD: $35.82$ years $\pm$ $11.57$, age range: $18–65$ years). The rs-fMRI preprocessing was performed in a standard pipeline using BRAinNetome fMRI Toolkit (Brant;\url{http://brant.brainnetome.org}), with the following steps: $1$) deletion of the first $10$ time points and slice-timing, $2$) realignment, $3$) co-registration of rs-fMRI scans into respective structural MRI scans, $4$) normalization to MNI standard space and re-sampling to $2\times2\times2$ mm$^3$ voxels, $5$) denoising with a multiple regression model and bandpass filtering (i.e., $0.01–0.08$ Hz), and $6$) smoothing images with $6$ mm full-width half-maximum (FWHM) Gaussian kernel. After pre-processing, we generated maps of the amplitude of time series (AM), (fractional) amplitude of low-frequency fluctuation (ALFF/fALFF) and regional homogeneity (ReHo) for each subject. To avoid the curse of dimensionality , we used a brain atlas (i.e., Neuromorphometrics) that comprises cortical and sub-cortical regions, and computed mean regional signals from AM, ALFF/fALFF, and ReHo maps (in total, $136$ features per subject). We additionally generated the functional connectivity (FC) between regions for each subject, resulting $8,911$ FC features per subject. Age and sex were also included in the prediction models.


 \begin{table}[h!]
    \centering
    \begin{tabular}{lcccc lccc}
    \cline{2-4} \cline{7-9}
        &Rank&AR&AC&& &Rank&AR&AC\\
        \cline{2-4} \cline{7-9}
Het-DRaF$^*$&$1$&$82.19$&$3.96$&&RaF&$6$&$81.36$&$5.61$\\
Het-RaF&$2$&$81.94$&$4.16$&&RaF-LDA&$7$&$81.36$&$4.78$\\
MPRaF-T&$3$&$81.65$&$5.06$&&RaF-PCA&$8$&$80.91$&$5.71$\\
MPRaF-P&$4$&$81.55$&$4.91$&&MPRaF-N&$9$&$80.55$&$5.75$\\
DRaF&$5$&$81.55$&$5.07$\\
 \cline{2-4} \cline{7-9}
    \end{tabular}
    \caption{ Comparison of baseline models with the proposed Het-DRaF model. (Here `$^*$' denotes the proposed algorithm introduced in this paper, AR stands for average rank and AC  stands for average accuracy)}
    \label{tab:average Friedman Rank}
\end{table}
 
 \begin{table}[h!]
\caption{Pairwise (count) win-tie-loss }
\label{tab:win-tie-loss}
\resizebox{\textwidth}{!}{
\begin{tabular}{llllllllll}
         & RaF            & MPRaF-T        & MPRaF-P        & MPRaF-N        & RaF-PCA        & RaF-LDA        & DRaF           & Het-RaF        & Het-DRaF \\ \hline
MPRaF-T  & {[}54,10,42{]} &                &                &                &                &                &                &                &          \\
MPRaF-P  & {[}56,17,33{]} & {[}50,13,43{]} &                &                &                &                &                &                &          \\ 
MPRaF-N  & {[}48,10,48{]} & {[}34,14,58{]} & {[}36,16,54{]} &                &                &                &                &                &          \\
RaF-PCA  & {[}45,15,46{]} & {[}40,8,58{]}  & {[}38,11,57{]} & {[}50,15,41{]} &                &                &                &                &          \\
RaF-LDA  & {[}57,10,39{]} & {[}49,10,47{]} & {[}49,14,43{]} & {[}62,13,31{]} & {[}61,17,28{]} &                &                &                &          \\
DRaF     & {[}56,14,36{]} & {[}52,8,46{]}  & {[}47,17,42{]} & {[}51,11,44{]} & {[}49,15,42{]} & {[}44,11,51{]} &                &                &          \\
Het-RaF  & {[}62,14,30{]} & {[}59,11,36{]} & {[}52,18,36{]} & {[}62,12,32{]} & {[}66,12,28{]} & {[}59,12,35{]} & {[}60,10,36{]} &                &          \\
Het-DRaF & {[}60,11,35{]} & {[}61,13,32{]} & {[}59,9,38{]}  & {[}67,11,28{]} & {[}67,16,23{]} & {[}59,14,33{]} & {[}61,12,33{]} & {[}50,14,42{]} &   \\\hline     
\end{tabular}}
\end{table}

\begin{table}[h!]
\caption{Significance difference of classification among the baseline and the proposed Het-DRaF models with Nemenyi post hoc tests (accuracy based).}
\label{tab:signifcdf}
\resizebox{\textwidth}{!}{
\begin{tabular}{lccccccccc}  \hline
         & RaF & MPRaF-T & MPRaF-P & MPRaF-N & RaF-PCA & RaF-LDA & DRaF & Het-RaF & Het-DRaF \\ \hline
RaF      &     &         &         &         &         &         &      & r-      & r-       \\
MPRaF-N  &     &         &         &         &         &         &      & r-      & r-       \\
RaF-PCA  &     &         &         &         &         &         &      & r-      & r-       \\
Het-RaF  & r+  &         &         & r+      & r+      &         &      &         &          \\
Het-DRaF & r+  &         &         & r+      & r+      &         &      &         &         \\ \hline
 \end{tabular}}
Here, r+ represents the row model which has better performance than the column model. r- represents the row model which has worse performance than the respective column model. Blank entities signify that the models do not differ significantly from one another.
\end{table}

\begin{table}[h!]
    \centering
    \resizebox{\columnwidth}{!}{
    \begin{tabular}{cccccccccc}
      \hline
Features & RaF & MPRaF-T & MPRaF-P & MPRaF-N & RaF-PCA & RaF-LDA & DRaF & Het-RaF & Het-DRaF \\ \hline
AM     & 67.77 & 68.44   & 65.05   & 67.04   & 66.44   & 70.47   & 69.14 & 64.31    & \textbf{73.91 }             \\
ALFF   & 65.12 & 63.02   & 65.03   & 62.96   & 69.18  & 67.83   & 62.25 & \textbf{71.9}     & 65.71              \\
fALFF  & 62.31 & 65.01   & 68.44   & 67.04   & 66.32   & 69.78   & \textbf{71.22} & 70.55    & 67.11              \\
ReHO   & \textbf{71.2}  & 70.57   & 69.76   & 67.13   & 70.57   & 71.18   & 65.7  & 68.56    & 69.86              \\
FC    & 63.58 & 68.43   & 63.58   & 68.39   & 64.33   & 62.27   & 65.03 & \textbf{72.46 }   & 71.17\\  
\hline
    \end{tabular}}
    Note: AM, the amplitude of time series; ALFF/fALFF , (fractional) amplitude of low-frequency fluctuation, ReHo, regional homogeneity; FC, functional connectivity.
    \caption{Accuracy of proposed and baselines models on Schizophrenia dataset classification followed by different functional metrics extracted from rs-fMRI brain scans}
    \label{table:classification-schizophrenia}
\end{table}
\subsection{Analysis of experimental results on Schizophrenia dataset}

In this study, we also examined the reliability of proposed classification model on a neuroimaging-based dataset. To this end, we extracted the functional brain features of $72$ Schizophrenia patients as well as $74$ healthy controls. We considered both low-dimensional brain features (i.e., extracted from AM, ALFF, fALFF and ReHo maps; in total, $136$ features per subject) as well as FC features which presents a high-dimensional feature set (in total, $8,911$ features per subject). Table \ref{table:classification-schizophrenia} shows the accuracy of different prediction models on various functional brain metrics extracted from rs-fMRI scans, whereas Table \ref{table:Rank-schizophrenia} displays ranking different prediction models per features. 
Regarding the low dimensional brain features sets, Het-DRaF model showed a superior result than baseline models in terms of AM features. For example, our prediction model exhibited an accuracy of $73.91\%$ on AM features, while the Het-RaF model showed an accuracy of $64.31\%$ on same data. Regarding other low-dimensional feature sets (i.e., ALFF, fALFF and ReHo), other baseline models showed a slightly better performance than proposed model. The Het-DRaF and Het-RaF models showed a superior performance (accuracy $>71\%$) than other models on FC features, indicating that proposed model is capable of delivering reliable results for classification studies based on high-dimensional neuroimaging features. The highest accuracy (i.e., $73.91\%$) on Schizophrenia dataset was achieved by proposed model with AM feature set.

\begin{table}[h!]
    \centering
    \resizebox{\columnwidth}{!}{
    \begin{tabular}{cccccccccc} \hline
     Features & RaF & MPRaF-T & MPRaF-P & MPRaF-N & RaF-PCA & RaF-LDA & DRaF & Het-RaF & Het-DRaF \\ \hline
AM    & 5   & 4   & 8   & 6 & 7   & 2 & 3 & 9 & \textbf{1} \\
ALFF  & 5   & 7   & 6   & 8 & 2   & 3 & 9 & \textbf{1} & 4 \\
fALFF & 9   & 8   & 4   & 6 & 7   & 3 & \textbf{1} & 2 & 5 \\
ReHO  & \textbf{1}   & 3.5 & 6   & 8 & 3.5 & 2 & 9 & 7 & 5 \\
FC    & 7.5 & 3   & 7.5 & 4 & 6   & 9 & 5 & \textbf{1} & 2 \\ \hline
    \end{tabular}}
    Note: The model predictions are ranked based on their performances per feature.
    \caption{Rank of proposed and baselines models on Schizophrenia dataset based on different functional metrics extracted from rs-fMRI brain scans.}
    \label{table:Rank-schizophrenia}
\end{table}
\begin{landscape}
\begin{footnotesize}
\begin{longtable}[t]{|p{0.27\textwidth}|p{0.06\textwidth}|p{0.09\textwidth}|p{0.09\textwidth}|p{0.092\textwidth}|p{0.092\textwidth}|p{0.092\textwidth}|c|c|c|p{0.092\textwidth}|}
\caption[this is]{Classification accuracy of baseline and Het-DRaF models }
\label{table:averageRaF} \\
\hline
Datasets &RaF \cite{breiman2001random}&	MPRaF-T \cite{zhang2014oblique}&	MPRaF-P \cite{zhang2014oblique}&	MPRaF-N \cite{zhang2014oblique}&RaF-PCA \cite{zhang2014random}&RaF-LDA \cite{zhang2014random}	&DRaF \cite{han2020double}	&Het-RaF \cite{katuwal2020heterogeneous} &	 Het-DRaF$^*$\\
\hline 
\endfirsthead
\multicolumn{10}{c}%
{{\bfseries \tablename\ \thetable{} -- continued from previous page}} \\
\hline
Datasets &RaF \cite{breiman2001random}&	MPRaF-T \cite{zhang2014oblique}&	MPRaF-P \cite{zhang2014oblique}&	MPRaF-N \cite{zhang2014oblique}&RaF-PCA \cite{zhang2014random}&RaF-LDA \cite{zhang2014random}	&DRaF \cite{han2020double}	&Het-RaF \cite{katuwal2020heterogeneous} &	 Het-DRaF$^*$\\
 \hline 
\endhead
\hline \multicolumn{10}{|r|}{{Continued on next page}} \\ \hline
\endfoot
\endlastfoot

\hline
  abalone&$64.68$&$64.99$&$65.54$&$65.06$&$64.85$&$65.4$&$64.18$&$64.7$&$64.89$\\
  balance-scale&$86.7$&$89.42$&$88.94$&$89.42$&$88.62$&$89.42$&$82.69$&$88.3$&$85.9$\\
  cardiotocography-10clases&$86.11$&$82.44$&$85.59$&$79.8$&$84.37$&$84.84$&$87.15$&$84.89$&$85.69$\\
acute-inflammation&$100$&$100$&$100$&$100$&$100$&$100$&$100$&$100$&$100$\\
balloons&$81.25$&$87.5$&$87.5$&$93.75$&$81.25$&$75$&$87.5$&$81.25$&$81.25$\\
acute-nephritis&$100$&$100$&$100$&$100$&$100$&$100$&$100$&$100$&$100$\\
blood&$76.6$&$76.74$&$77.27$&$77.81$&$76.6$&$77.01$&$75.67$&$76.87$&$76.47$\\
annealing&$54.25$&$76$&$38.75$&$76$&$62.25$&$65$&$37$&$63.75$&$60.5$\\
breast-cancer&$73.94$&$73.94$&$73.94$&$73.94$&$76.06$&$76.76$&$75.35$&$73.24$&$75.35$\\
breast-cancer-wisc&$97.29$&$97.71$&$97.43$&$97$&$97.14$&$97.43$&$97.14$&$97.57$&$97.71$\\
arrhythmia&$73.01$&$63.05$&$73.23$&$61.5$&$65.49$&$67.04$&$73.67$&$71.02$&$71.02$\\
audiology-std&$75$&$70$&$76$&$24$&$55$&$48$&$78$&$68$&$72$\\
car&$96.93$&$95.31$&$96.99$&$88.08$&$96.76$&$96.76$&$97.05$&$97.86$&$98.03$\\
cardiotocography-3clases&$94.02$&$92.75$&$94.26$&$91.57$&$92.33$&$93.27$&$94.92$&$92.75$&$93.55$\\
conn-bench-sonar-mines-rocks&$76.92$&$78.37$&$78.85$&$78.37$&$76.44$&$78.85$&$79.33$&$81.25$&$80.77$\\

breast-cancer-wisc-diag&$95.6$&$96.83$&$96.83$&$97.71$&$95.6$&$97.01$&$95.77$&$96.83$&$97.01$\\
breast-cancer-wisc-prog&$80.1$&$80.61$&$79.59$&$81.12$&$80.1$&$80.61$&$81.63$&$80.61$&$82.14$\\
breast-tissue&$70.19$&$69.23$&$71.15$&$71.15$&$73.08$&$75$&$73.08$&$71.15$&$72.12$\\
ecoli&$86.31$&$87.2$&$88.99$&$87.8$&$87.5$&$87.5$&$88.1$&$87.5$&$87.5$\\
energy-y1&$94.79$&$92.58$&$94.92$&$94.14$&$94.53$&$95.7$&$95.83$&$95.7$&$96.09$\\

congressional-voting&$62.39$&$61.24$&$61.01$&$61.24$&$60.55$&$61.01$&$61.7$&$61.24$&$61.47$\\

conn-bench-vowel-deterding&$98.48$&$99.78$&$99.13$&$99.62$&$99.57$&$99.46$&$98.97$&$99.84$&$100$\\
credit-approval&$87.5$&$86.92$&$86.05$&$88.08$&$88.08$&$87.5$&$87.5$&$87.35$&$87.35$\\
cylinder-bands&$81.25$&$76.17$&$80.47$&$73.05$&$78.71$&$77.73$&$82.03$&$79.3$&$80.27$\\
dermatology&$98.35$&$98.08$&$98.35$&$96.7$&$97.8$&$97.8$&$98.08$&$98.35$&$97.25$\\
echocardiogram&$84.85$&$84.85$&$85.61$&$84.09$&$84.09$&$83.33$&$84.09$&$84.09$&$84.09$\\
fertility&$88$&$89$&$89$&$88$&$88$&$88$&$88$&$89$&$88$\\
energy-y2&$89.06$&$89.45$&$89.84$&$89.32$&$89.97$&$89.71$&$88.28$&$88.54$&$89.71$\\
heart-cleveland&$57.89$&$61.51$&$57.57$&$59.21$&$58.22$&$59.21$&$55.59$&$59.21$&$59.21$\\
heart-hungarian&$83.9$&$84.93$&$84.25$&$84.25$&$84.59$&$84.59$&$84.25$&$84.25$&$83.9$\\
flags&$67.19$&$55.21$&$64.58$&$56.77$&$56.25$&$57.29$&$66.67$&$59.9$&$58.85$\\
glass&$73.11$&$69.34$&$75.47$&$70.28$&$70.75$&$72.17$&$76.89$&$70.28$&$71.23$\\
haberman-survival&$71.05$&$71.71$&$71.05$&$72.37$&$70.07$&$71.38$&$69.74$&$70.39$&$67.43$\\
hayes-roth&$87.5$&$86.61$&$84.82$&$81.25$&$89.29$&$87.5$&$89.29$&$88.39$&$85.71$\\
hepatitis&$83.33$&$82.05$&$82.05$&$86.54$&$82.69$&$84.62$&$82.69$&$83.33$&$85.9$\\
hill-valley&$53.84$&$66.75$&$63$&$65.88$&$64.03$&$66.25$&$54.17$&$66.3$&$67.9$\\
heart-switzerland&$41.13$&$43.55$&$41.13$&$44.35$&$43.55$&$45.16$&$41.94$&$44.35$&$45.16$\\
heart-va&$35.5$&$34.5$&$35.5$&$36.5$&$34$&$37.5$&$36$&$32.5$&$34.5$\\
ionosphere&$91.76$&$93.75$&$93.47$&$93.18$&$94.03$&$94.6$&$91.48$&$93.47$&$94.03$\\
iris&$95.27$&$97.3$&$97.3$&$97.97$&$95.95$&$96.62$&$95.95$&$96.62$&$95.95$\\
horse-colic&$86.4$&$86.03$&$87.87$&$87.5$&$82.35$&$85.29$&$86.76$&$87.13$&$85.29$\\
ilpd-indian-liver&$71.4$&$70.72$&$71.23$&$71.23$&$73.29$&$71.23$&$71.23$&$72.95$&$72.6$\\
image-segmentation&$93.8$&$94.18$&$94.75$&$92.46$&$94.96$&$95.07$&$94.85$&$95.51$&$95.64$\\
libras&$76.94$&$84.17$&$79.17$&$79.44$&$80.28$&$81.39$&$79.72$&$87.22$&$85.28$\\
low-res-spect&$90.79$&$91.17$&$91.35$&$89.47$&$90.6$&$91.54$&$91.54$&$91.35$&$92.11$\\
led-display&$74.3$&$72$&$73.7$&$72.4$&$73.9$&$73.6$&$71.7$&$72.5$&$71.8$\\
lenses&$83.33$&$79.17$&$83.33$&$79.17$&$79.17$&$87.5$&$83.33$&$83.33$&$79.17$\\
mammographic&$81.98$&$81.67$&$80.63$&$81.67$&$80.63$&$80.42$&$79.9$&$80.31$&$80$\\
molec-biol-promoter&$84.62$&$79.81$&$84.62$&$82.69$&$71.15$&$78.85$&$91.35$&$84.62$&$87.5$\\
lung-cancer&$46.88$&$46.88$&$50$&$53.13$&$40.63$&$43.75$&$50$&$46.88$&$50$\\
lymphography&$79.05$&$85.14$&$83.11$&$83.78$&$84.46$&$84.46$&$85.81$&$88.51$&$84.46$\\
monks-2&$66.78$&$66.9$&$66.9$&$67.01$&$66.84$&$67.01$&$66.55$&$66.72$&$66.67$\\
monks-3&$53.01$&$56.6$&$52.78$&$54.34$&$53.36$&$52.89$&$52.78$&$54.63$&$57.52$\\
molec-biol-splice&$94.2$&$86.57$&$93.1$&$85.01$&$84.1$&$89.9$&$94.7$&$91.15$&$91.06$\\
monks-1&$59.95$&$60.59$&$58.39$&$57.52$&$58.04$&$58.16$&$60.65$&$58.45$&$58.39$\\
musk-2&$97.21$&$96.12$&$95.94$&$95.69$&$95.98$&$96.12$&$98.12$&$96.62$&$97.1$\\
OM\_nucleus\_4d&$77.65$&$81.57$&$82.55$&$80.69$&$82.55$&$82.65$&$79.8$&$84.41$&$83.92$\\
mushroom&$100$&$100$&$100$&$100$&$100$&$100$&$100$&$100$&$100$\\
musk-1&$86.13$&$86.97$&$83.82$&$87.18$&$86.13$&$83.82$&$86.34$&$84.66$&$87.82$\\
OT\_states\_5b&$90.9$&$92.54$&$92$&$92.65$&$92.65$&$93.31$&$92.21$&$93.53$&$93.42$\\
optical&$96.08$&$95.66$&$96.26$&$84.65$&$95.72$&$91.62$&$96.91$&$96.52$&$96.97$\\
OM\_states\_2f&$91.37$&$91.57$&$92.16$&$91.86$&$91.86$&$92.16$&$92.35$&$93.04$&$92.94$\\
OT\_nucleus\_2f&$79.39$&$81.58$&$82.79$&$83.11$&$82.57$&$82.24$&$80.7$&$83$&$83.44$\\
pima&$76.69$&$75.26$&$75.52$&$75.13$&$74.61$&$74.87$&$73.96$&$75$&$76.56$\\
pittsburg-bridges-MATERIAL&$91.35$&$93.27$&$91.35$&$92.31$&$92.31$&$92.31$&$88.46$&$91.35$&$90.38$\\
ozone&$97.08$&$97.2$&$97.16$&$97.16$&$97.16$&$97.16$&$97.08$&$97.12$&$97.12$\\
page-blocks&$97.08$&$96.98$&$97.3$&$96.78$&$97.09$&$97.13$&$97.08$&$97.26$&$97.22$\\
parkinsons&$88.78$&$92.35$&$89.8$&$91.84$&$87.76$&$90.82$&$90.31$&$89.8$&$89.29$\\
pittsburg-bridges-T-OR-D&$88$&$88$&$88$&$88$&$88$&$90$&$89$&$88$&$88$\\
pittsburg-bridges-TYPE&$68.27$&$69.23$&$67.31$&$66.35$&$71.15$&$66.35$&$67.31$&$70.19$&$71.15$\\
pittsburg-bridges-REL-L&$74.04$&$75.96$&$73.08$&$75.96$&$75$&$73.08$&$73.08$&$75$&$75.96$\\
pittsburg-bridges-SPAN&$61.96$&$72.83$&$63.04$&$67.39$&$69.57$&$71.74$&$60.87$&$64.13$&$67.39$\\
ringnorm&$95.19$&$90.41$&$90.81$&$90.85$&$97.01$&$97.09$&$95.46$&$97.26$&$97.66$\\
seeds&$93.27$&$94.71$&$91.83$&$91.83$&$93.75$&$92.31$&$93.75$&$93.75$&$94.71$\\
planning&$70$&$67.78$&$70$&$70$&$70.56$&$69.44$&$69.44$&$71.11$&$71.67$\\
plant-texture&$77.94$&$77.25$&$76.06$&$75.06$&$75.81$&$76.81$&$80.56$&$80.63$&$81.88$\\
post-operative&$72.73$&$71.59$&$70.45$&$71.59$&$69.32$&$67.05$&$70.45$&$67.05$&$70.45$\\
primary-tumor&$54.88$&$51.83$&$55.18$&$52.44$&$53.05$&$56.1$&$54.88$&$55.49$&$53.35$\\
spect&$68.95$&$61.56$&$65.46$&$59.68$&$60.75$&$61.29$&$65.05$&$64.25$&$61.83$\\
spectf&$91.98$&$91.98$&$91.98$&$91.98$&$91.98$&$91.84$&$91.98$&$86.36$&$91.98$\\
soybean&$90.29$&$89.23$&$90.56$&$82.71$&$89.83$&$86.3$&$90.36$&$91.09$&$91.09$\\
spambase&$94.39$&$94.5$&$94.15$&$94.11$&$94.8$&$94.48$&$94.72$&$95.13$&$95.35$\\
statlog-heart&$85.45$&$87.31$&$86.19$&$86.19$&$85.45$&$85.45$&$85.07$&$86.94$&$85.07$\\
statlog-image&$97.27$&$97.66$&$97.57$&$96.66$&$97.88$&$97.92$&$97.88$&$97.75$&$98.35$\\
statlog-australian-credit&$67.3$&$65.26$&$66.57$&$67.15$&$63.66$&$63.23$&$64.39$&$65.84$&$64.1$\\
statlog-german-credit&$77.5$&$74.8$&$75.4$&$73.9$&$75.3$&$77.7$&$77.4$&$76.2$&$76.3$\\
synthetic-control&$97.67$&$99.83$&$98.5$&$98.33$&$97.17$&$99.17$&$98.5$&$99.5$&$99$\\
teaching&$59.21$&$58.55$&$60.53$&$57.24$&$55.92$&$60.53$&$58.55$&$62.5$&$61.18$\\
statlog-landsat&$89.94$&$89.99$&$89.99$&$89.04$&$89.78$&$89.88$&$90.78$&$90.61$&$90.79$\\
statlog-vehicle&$73.58$&$76.3$&$77.73$&$75.95$&$78.08$&$79.03$&$75.71$&$76.78$&$78.67$\\
steel-plates&$78.04$&$78.04$&$76.75$&$75.15$&$75.05$&$76.49$&$78.4$&$77.63$&$77.27$\\
thyroid&$98.88$&$95.86$&$98.89$&$93.26$&$98.7$&$97.65$&$98.96$&$98.27$&$98.53$\\
tic-tac-toe&$97.91$&$97.49$&$97.7$&$94.77$&$97.07$&$98.01$&$98.64$&$99.06$&$98.95$\\
twonorm&$96.8$&$97.59$&$97.68$&$97.57$&$97.68$&$97.55$&$96.8$&$97.49$&$97.54$\\
vertebral-column-2clases&$83.77$&$86.69$&$86.04$&$86.04$&$85.06$&$86.69$&$82.14$&$84.42$&$83.77$\\
titanic&$78.95$&$78.68$&$78.95$&$78.32$&$78.95$&$78.95$&$78.95$&$78.95$&$78.95$\\
trains&$87.5$&$100$&$87.5$&$87.5$&$87.5$&$87.5$&$87.5$&$87.5$&$100$\\
waveform&$84.54$&$85.4$&$85.04$&$85.44$&$84.8$&$85.4$&$83.76$&$85.98$&$85.38$\\
waveform-noise&$85.5$&$85.2$&$86.24$&$85.74$&$85.08$&$85.84$&$85.22$&$86.36$&$86.26$\\
vertebral-column-3clases&$83.44$&$84.09$&$83.44$&$83.77$&$83.77$&$83.77$&$84.74$&$84.74$&$87.01$\\
wall-following&$99.3$&$94.24$&$98.41$&$93.71$&$96.17$&$96.19$&$99.52$&$96.52$&$96.79$\\
wine-quality-red&$65.81$&$68$&$67.38$&$68.19$&$68.31$&$67.56$&$68$&$69.56$&$68.38$\\
wine-quality-white&$67.01$&$67.28$&$67.57$&$66.14$&$67.87$&$67.69$&$68.2$&$68.85$&$68.34$\\
wine&$97.73$&$98.86$&$99.43$&$97.73$&$97.16$&$98.86$&$97.73$&$99.43$&$98.86$\\
yeast&$61.52$&$62.06$&$61.79$&$62.2$&$62.53$&$62.53$&$60.58$&$62.26$&$61.86$\\
zoo&$99$&$99$&$98$&$98$&$99$&$99$&$98$&$99$&$99$\\ 
\hline
\multicolumn{10}{l}{Here, $^*$ denotes the proposed method.}\\
\multicolumn{10}{l}{OM denotes oocytes\_merluccius, OT denotes oocytes\_trisopterus.}
    \end{longtable}
    \end{footnotesize}
\end{landscape}

\section{Conclusion and Future Direction}
\label{sec:6}
In this paper, we proposed heterogeneous oblique double random forest for the classification problems. The proposed model uses several linear classifiers at non-leaf nodes based on the bootstrapped data to split the data. The decision hyperplane which optimizes the impurity criteria is selected for partitioning the input data instead of the bootstrapped data. This strategy leads to more unique samples at each non-leaf node which results in better generalization performance. 
The proposed Het-DRaF model uses bootstrapping on the non-leaf data unlike the standard RaF and heterogeneous oblique RaF which follow bootstrapping at the root node only. Moreover, the standard RaF and DRaF models split the data at each node based on univariate features while as the proposed Het-DRaF model generates the optimal splits based on multivariate features which results in capturing the  geometric features of the data. The standard oblique decision trees based ensembles (MPRaF-P, MPRaF-T, MPRaF-N) which use a single decision hyperplane on non-leaf nodes based on the bootstrapped data, the proposed Het-DRaF model uses multiple linear classifiers to generate the best split based on the original data rather than bootstrapped data.
We evaluated the performance of several existing models and the proposed Het-DRaF model on the benchmark UCI repository datasets. Moreover, we used the classification models for the diagnosis of the Schizophrenia disease. Experimental results demonstrate that the performance of the proposed model is better compared to the baseline models. Similar to Het-RaF model, the proposed model evaluates multiple linear classifiers, hence, the proposed Het-DRaF model involves more computations compared to other baseline models. In future, one can explore more  approaches for generating the more diversified heterogeneous decision trees. In addition to classification, one can explore this work in the area of regression and time series forecasting. One can benchmark the performance of the standard random forest based models, double random forest based models, standard heterogeneous random forest, proposed heterogeneous double random forest and XGBoost to see the performance which can guide in choosing the best model for the target applications. 

\section*{Acknowledgment}
This research is supported by SERB, a statutory body (Indian Government) covered by Ramanujan Fellowship scheme, (file No. SB/S2/RJN-001/2016), and DST (Govt. of India) covered by Interdisciplinary Cyber Physical Systems program (file no. DST/ICPS/CPS-Individual/2018/276). The grateful acknowledgement is given to the IIT Indore, India for providing facilities and support.
\bibliographystyle{IEEEtranN}
\bibliography{refs.bib}
\end{document}